\documentclass{article}
\usepackage{spconf,amsmath,amssymb,graphicx,booktabs}
\usepackage[hyphens]{url}
\usepackage[hidelinks]{hyperref}
\hypersetup{pdftitle={JevSoup: System-One Routing for Training-Free LoRA Composition},
  pdfsubject={Working draft for author revision; completed full benchmark and ablations}}
\graphicspath{{figures/}}
\newcommand{\method}{JevSoup}

\title{JevSoup: System-One Routing for\\Training-Free LoRA Composition}
\name{\shortstack{
Xiuying Wang$^{1,\dagger}$, Jiahua Cheng$^{1,\dagger}$, Shuotian Li$^{2,\dagger}$, Yufan Cheng$^{3}$,\\
Bowen Deng$^{1}$, Zhexuan Bai$^{1}$, Yichen Li$^{4,*}$
}\thanks{$^{\dagger}$These authors contributed equally. $^{*}$Corresponding author.}}
\address{
$^{1}$Beijing University of Posts and Telecommunications\quad $^{2}$University of Malaya\\
$^{3}$Georgian College\quad $^{4}$Huazhong University of Science and Technology
}

\begin{document}
\maketitle
\begin{abstract}
Building adaptable AI systems requires effective coordination of specialized
capabilities across diverse tasks. Low-rank adaptation (LoRA) enables modular
expertise, but existing routing approaches may require auxiliary data,
additional training, or autoregressive decoding. We propose \method, a
training-free framework separating System~One expert routing from System~Two
execution. Using only the input and expert descriptions, Jev selects two
experts through structured probabilities. \method{} retains the leading
expert's update, projects the second onto the orthogonal complement of the
first update's row space, and combines them with equal weights. Across 14
PorTAL tasks and three Qwen3 scales, \method{} achieves absolute gains of up
to 1.19\% in task-macro and 1.21\% in sample-micro accuracy over the strongest
evaluated external baselines. Our code is available at \url{https://github.com/Leowang980/JevSoup}.

\end{abstract}

\begin{figure*}[t]
  \centering
  \includegraphics[width=\textwidth]{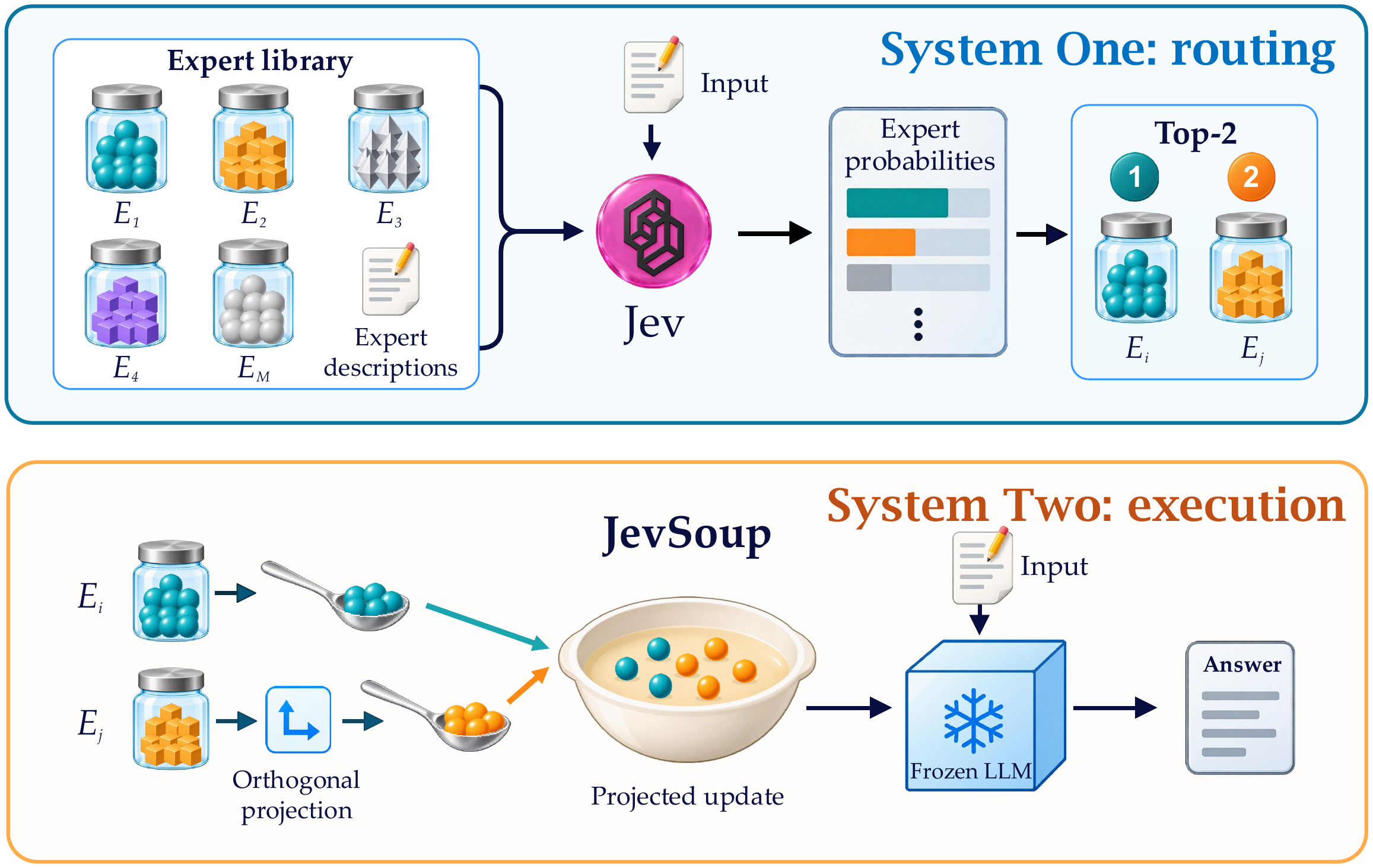}
  \caption{Overview of \method. Top (System One): Jev uses the input
  and expert descriptions to rank the experts and select an ordered Top-2
  pair. Bottom (System Two): The second LoRA update is orthogonally
  projected relative to the first, and the two complete updates are equally
  combined in the frozen language model to produce the answer. Probability
  bars are schematic.}
  \label{fig:framework}
\end{figure*}

\section{Introduction}
\label{sec:intro}

System~One and System~Two describe complementary aspects of cognition:
judgments that arise with little conscious effort and reasoning that requires
sustained attention~\cite{kahneman2011thinking}. This distinction offers an
analogy for how AI assistants coordinate specialized capabilities.
A customer-support assistant, for example, may need billing, product-policy,
or troubleshooting expertise depending on the request~\cite{shi2026tau}.
Low-rank adaptation
(LoRA)~\cite{hu2021lora} provides a practical way to make such expertise modular,
representing specialized capabilities as lightweight adapters to a shared
language model. As these adapters accumulate into an expert library, using
them effectively requires deciding which experts are relevant to each input
and how their contributions should be combined.

To make effective use of these expert libraries, prior work has explored
a range of routing and composition strategies.
MoLE~\cite{wu2024mixture} and related gated mixtures~\cite{xu2025meteora}
train expert selection networks, while
LoraRetriever~\cite{zhao2024loraretriever} learns an
instruction-tuned retriever.
Without router training, methods can still use auxiliary data to select
components for averaging,
as in Model Soups~\cite{wortsman2022model} and
AdapterSoup~\cite{chronopoulou2023adaptersoup}, build representations for
retrieval~\cite{han2025hilora,dhasade2026effective},
or optimize mixture weights without gradients, as in
LoraHub~\cite{huang2023lorahub}.

However, representative examples may be unavailable, and learned routers
require additional training and may need updating as the expert library
evolves. LoGo~\cite{lee2026lora} and Arrow~\cite{ostapenko2024towards}
avoid these dependencies using adapter activations and prototypes derived from
adapter weights, respectively. These signals tie routing to the backbone and
leave task relevance implicit. LLMs can instead match inputs to explicit
expert descriptions~\cite{shekar2025adaptive}. This approach still requires
autoregressive decoding and parsing.
Jev~\cite{almeida2026jev}, a pretrained System~One model, offers a structured
alternative by mapping the input and expert descriptions directly to
probabilities over expert identifiers.
Yet selection alone does not determine how to combine the chosen adapters.
Routing probabilities need not be suitable mixture weights, and direct
averaging does not explicitly control overlap between parameter updates.
Inspired by the use of orthogonal projection to limit interference in
continual learning~\cite{farajtabar2020orthogonal}, we treat composition as a
separate problem from expert selection.

To this end, we propose \method, a training-free framework that separates
System~One expert selection from System~Two task execution. Jev selects an
ordered pair of experts using only the input and their capability descriptions.
For composition, \method{} retains the higher-ranked update and projects the
second onto the orthogonal complement of the first update's row space.
The resulting updates are combined with equal weights in a single frozen
backbone. This requires neither expert training samples nor additional
router or expert training.

Our main contributions are threefold:
\begin{itemize}
  \item We formulate LoRA-based inference as System~One routing and
  System~Two execution, identifying dependence on training samples, router
  retraining, and autoregressive decoding as limitations of existing
  routing approaches.
  \item We propose \method, a training-free framework that combines
  Jev-based structured selection with an asymmetric composition rule:
  retain the higher-ranked update, orthogonalize the second against its
  row space, and fuse them with equal weights.
  \item Across 14 tasks and three model scales, \method{} achieves absolute
  gains of up to 1.19\% in task-macro and 1.21\% in sample-micro accuracy
  over the strongest evaluated external baselines.
\end{itemize}

\section{Methodology}
\label{sec:method}

\subsection{Problem Definition}
\label{sec:library}

Let $f_{\theta_0}$ be a frozen language model. We consider a library of $M$
compatible LoRA experts, $\mathcal E=\{E_1,E_2,\ldots,E_M\}$.
All experts have frozen LoRA parameters and adapt the same weight matrices
of the backbone. Each expert $E_i$ is associated with an expert card $c_i$,
containing its identifier and a natural-language description of its
capabilities.

For an adapted weight matrix $W_0$, expert $E_i$ contributes the complete
LoRA~\cite{hu2021lora} update
\begin{equation}
  \Delta W_i=s_iB_iA_i,
  \qquad s_i=\alpha_i/r_i,
  \label{eq:lora}
\end{equation}
where $A_i\in\mathbb R^{r_i\times d_{\rm in}}$ and
$B_i\in\mathbb R^{d_{\rm out}\times r_i}$ are frozen factors,
and $r_i$ and $\alpha_i$ denote the rank and scaling hyperparameter.
We omit matrix indices; the same definitions apply independently at each
adapted matrix.

Given an input $x$ and the expert library $\mathcal E$, our goal is to select
relevant experts and compose their updates into an input-dependent update
$\Delta W(x)$, yielding adapted weights $W_0+\Delta W(x)$ for task execution.
In our multiple-choice evaluation, $x$ contains the prompt and all candidate
answers.
Gold answers and task labels are unavailable to routing.

\subsection{JevSoup Framework}
\label{sec:framework}

\method{} implements this selection--composition pipeline in two stages.
System One routing identifies an ordered pair of experts from their
capability descriptions. System Two execution composes their updates and
uses them in a single frozen backbone. The framework is illustrated in
Fig.~\ref{fig:framework}.

\smallskip
\noindent\textbf{System One routing.}
\label{sec:selection}
Expert selection can be expressed as matching the input to the
capabilities described in the expert cards. We use
Jev~\cite{almeida2026jev}, a pretrained System One model, to make this
decision through a typed choice interface rather than generate a textual
expert name. Jev is instructed to select the most suitable expert, not
solve the task. Conditioned on the fixed cards, the router $R_\phi$ returns
a probability vector over expert identifiers:
\begin{equation}
  \mathbf p(x)=R_{\phi}(x),\qquad
  p_i(x)\geq 0,\quad \sum_{i=1}^{M}p_i(x)=1.
  \label{eq:router}
\end{equation}
After validating the response and normalizing its probabilities, we select
the ordered Top-2 pair $(a,b)$, with $p_a(x)\geq p_b(x)$, breaking ties by
expert identifier. This decision is made once per input and shared across
all adapted matrices and candidate-answer evaluations.

\smallskip
\noindent\textbf{System Two execution.}
\label{sec:orthogonal}
In continual learning, Orthogonal Gradient Descent
(OGD)~\cite{farajtabar2020orthogonal} addresses interference by projecting
new-task loss gradients away from the span of stored gradients of past-task
model outputs.
Inspired by this principle, we retain one expert update and remove the
component of the other that overlaps its row space. Unlike OGD, this
operation acts on frozen LoRA updates without computing gradients or
performing training.

The routing order determines which update to retain: the higher-ranked
expert $a$ serves as the anchor, and only expert $b$ is projected.
We omit the input dependence below. Let $k_a=\operatorname{rank}(\Delta W_a)$
and $Q_a\in\mathbb R^{d_{\rm in}\times k_a}$ have orthonormal columns
spanning the row space of the complete update $\Delta W_a$.
Then $Q_aQ_a^\top$ projects onto input-space directions used by this update.
Let $I$ denote the $d_{\rm in}\times d_{\rm in}$ identity matrix.
With projection strength $\lambda\in[0,1]$, the second update becomes
\begin{equation}
  \widetilde{\Delta W}_b(\lambda)
  =\Delta W_b\bigl(I-\lambda Q_aQ_a^\top\bigr),
  \label{eq:projection}
\end{equation}
We then combine the anchor and projected update with equal weights:
\begin{equation}
  \Delta W(\lambda)
  =\tfrac12\bigl(\Delta W_a+\widetilde{\Delta W}_b(\lambda)\bigr).
  \label{eq:orthogonal}
\end{equation}
Equal weighting separates expert ranking from mixture magnitude without
learning additional coefficients. Each update retains its original LoRA
scaling. We do not restore the projected update's norm or renormalize the
branch weights, so projection can change both direction and magnitude.
The composed update is applied through two low-rank branches within one
backbone, not by averaging the $A$ and $B$ factors independently or
ensembling separate model predictions.

To implement this composition without forming dense update matrices,
we apply the projection directly to the second expert's low-rank factor:
\begin{equation}
  \widetilde A_b(\lambda)
  =A_b-\lambda(A_bQ_a)Q_a^\top,
  \label{eq:lowrank-projection}
\end{equation}
with $B_b$ and the original scale unchanged. We obtain $Q_a$ from the leading
$k_a$ right singular vectors of $R_aA_a$, where $R_a$ is the triangular factor
of the reduced QR decomposition of $B_a$.

\section{Experiments}
\label{sec:experiments}

\subsection{Experiment Setup}

\noindent\textbf{Benchmark.}
PorTAL~\cite{geistportal} provides task-specialized LoRA experts covering language understanding,
question answering, and commonsense reasoning. We evaluate on 19,477
multiple-choice examples across its 14 tasks and report task-macro and
sample-micro accuracy.
Our evaluation set is drawn from PorTAL's released validation split,
with development prompts excluded. The main comparison, ablations, and
sensitivity analysis use the same evaluation examples.

\smallskip
\noindent\textbf{Baseline.}
\textbf{Base} is the unadapted backbone.
\textbf{Adaptive Minds (AM)}~\cite{shekar2025adaptive} is adapted to select
two experts with the unadapted Qwen3 backbone; a keyword fallback fills
missing IDs, and their complete LoRA updates are equally weighted.
\textbf{LoGo}~\cite{lee2026lora} ranks and weights experts by last-block
query-projection update norms.
Our \textbf{AdapterSoup}~\cite{chronopoulou2023adaptersoup} variant retrieves
with Qwen3-Embedding-0.6B~\cite{zhang2025qwen3} using 100 training examples
per expert after excluding evaluation overlaps, then averages LoRA factors.
\textbf{Arrow}~\cite{ostapenko2024towards} routes with singular-vector
prototypes and combines LoRA factors with softmax weights.
All routing methods use Top-2.
All baselines share our backbone, expert library, and scoring protocol;
these are task adaptations, not replications of the original absolute scores.

\smallskip
\noindent\textbf{Configuration.}
We use Qwen3-1.7B, 4B, and 8B~\cite{yang2025qwen3} with 14 frozen PorTAL experts per backbone.
LoRA adapts query/value projections with rank 8 and $\alpha=16$.
We construct description-only expert cards from task definitions, with no
training examples, and use the same cards for Jev and AM.
We use Jev version \texttt{jev-1.13.0}\footnote{\url{https://docs.typesafe.ai/models}}
with card ordering fixed by seed 42.
Unless otherwise specified, we set the projection strength $\lambda$ to 1.
Inference uses BF16, seed 42, a 768-token prompt limit, and one candidate per
forward pass. Answers are scored by character-normalized continuation
log-probability. Current experiments run on an NVIDIA RTX PRO 6000 GPU.

\begin{table*}[t]
 \centering
 \caption{Main comparison on PorTAL (accuracy, \%). Macro and Micro denote
 task- and sample-averaged accuracy. Best and second-best values in each
 column are bold and underlined, respectively. At each model scale, all routing
 methods share the same backbone, expert library, and evaluation protocol.}
 \label{tab:main}
\begin{tabular*}{\textwidth}{@{\extracolsep{\fill}}lrrrrrr@{}}
\toprule
 & \multicolumn{2}{c}{Qwen3-1.7B} & \multicolumn{2}{c}{Qwen3-4B} & \multicolumn{2}{c}{Qwen3-8B} \\
\cmidrule(lr){2-3}\cmidrule(lr){4-5}\cmidrule(l){6-7}
Method & Macro & Micro & Macro & Micro & Macro & Micro \\
\midrule
Base & 58.05 & 61.79 & 64.05 & 69.76 & 68.15 & 74.41 \\
Adaptive Minds~\cite{shekar2025adaptive} & \underline{69.79} & \underline{67.15} & \underline{74.11} & 72.49 & \underline{77.07} & \underline{78.54} \\
LoGo~\cite{lee2026lora} & 68.74 & 64.96 & 71.43 & 69.75 & 75.94 & 77.01 \\
AdapterSoup~\cite{chronopoulou2023adaptersoup} & 69.18 & 66.85 & 73.61 & \underline{72.60} & 76.86 & 77.87 \\
Arrow~\cite{ostapenko2024towards} & 69.12 & 65.81 & 72.65 & 71.27 & 76.45 & 77.11 \\
\midrule
\textbf{JevSoup} & \textbf{70.91} & \textbf{67.40} & \textbf{75.30} & \textbf{73.81} & \textbf{77.29} & \textbf{78.73} \\
\bottomrule
\end{tabular*}

\end{table*}

\subsection{Main Results}

Among completed results in Table~\ref{tab:main}, \method{} has the highest
task-macro and sample-micro accuracy at all three scales. Relative to Base,
it gains 12.86, 11.25, and 9.14 macro points at 1.7B, 4B, and 8B.
AM Top-2 is the strongest external method in macro accuracy at all three
scales; \method{} exceeds it by 1.13, 1.19, and 0.21 points, respectively.
Its micro accuracy is also higher than each external baseline at all three
scales. The Jev-only control is examined alongside AM Top-2 in
Table~\ref{tab:ablation} to separate routing from orthogonal composition.

\subsection{Ablation Study}

Table~\ref{tab:ablation} examines seven 4B configurations: Random Top-2,
AM Top-2, Jev Top-1, and four Jev Top-2 variants with equal or probability
weights, each with or without projection. AM Top-2 and unprojected Jev
Top-2 use the same equal-weight composition, isolating expert selection.
The equal-weight projected variant is \method{}. All use the same backbone,
expert pool, and scoring protocol with model seed 42; Random Top-2 averages
over routing seeds 42--44.

\begin{table}[t]
 \centering
 \caption{4B ablation (accuracy, \%). Random Top-2 reports mean $\pm$
 standard deviation over three routing seeds. Prob. denotes routing
 probabilities renormalized over the selected pair. Proj. indicates whether
 full projection ($\lambda=1$) is applied to the second expert's update.}
 \label{tab:ablation}
\begingroup
\setlength{\tabcolsep}{3pt}
\begin{tabular*}{\columnwidth}{@{\extracolsep{\fill}}lcccc@{}}
\toprule
Selection & Weights & Proj. & Macro & Micro \\
\midrule
Random Top-2 & Equal & No & 72.43 $\pm$ 0.34 & 71.77 $\pm$ 0.16 \\
Jev Top-1 & Single & -- & 73.24 & 71.83 \\
Jev Top-2 & Prob. & No & 74.45 & 72.33 \\
Jev Top-2 & Prob. & Yes & \underline{74.86} & 72.41 \\
AM Top-2 & Equal & No & 74.11 & 72.49 \\
Jev Top-2 & Equal & No & 74.64 & \underline{73.49} \\
\midrule
\textbf{JevSoup} & Equal & Yes & \textbf{75.30} & \textbf{73.81} \\
\bottomrule
\end{tabular*}
\endgroup

\end{table}

Under equal weights and without projection, Jev Top-2 exceeds Random Top-2
by 2.21 macro and 1.72 micro points, and AM Top-2 by 0.53 and 1.00 points.
These comparisons support Jev-based selection under matched Top-2 fusion.
Jev Top-2 also exceeds Jev Top-1 by 1.40 macro and 1.66 micro points.
Equal weights outperform probability weights both with and without projection,
especially on micro accuracy. Overall, \method{} achieves the highest macro
(75.30\%) and micro (73.81\%) accuracy, 0.66 and 0.32 points above the
unprojected equal-weight Jev Top-2 variant.
Together, these results support the combination of Jev-based Top-2 selection,
equal weighting, and ordered projection in \method{}.

\subsection{Sensitivity Analysis}
\label{sec:sensitivity}

\begin{figure}[t]
 \centering
 \includegraphics[width=\columnwidth]{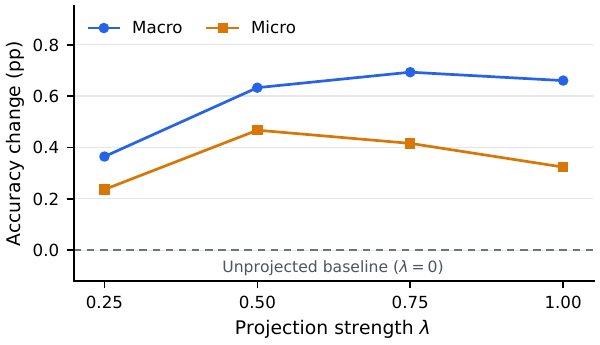}
 \caption{4B sensitivity to projection strength.
 Point-estimate accuracy changes relative to the unprojected control
 ($\lambda=0$, dashed line), with fixed routing and equal weights.}
 \label{fig:sensitivity}
\end{figure}

Figure~\ref{fig:sensitivity} varies $\lambda\in\{0.25,0.5,0.75,1\}$
with fixed routing and equal weights. All four settings yield positive
point-estimate gains over $\lambda=0$. Macro accuracy spans
75.01--75.34\% and micro accuracy 73.72--73.95\%, varying by 0.33 and
0.23 percentage points, respectively, on this evaluation set.
These small variations suggest that \method{} is relatively insensitive to
projection strength over the tested range.
We retain $\lambda=1$ for the main results.

\section{Conclusion}
\label{sec:conclusion}

We presented \method{}, a training-free framework that separates System One
expert routing from System Two task execution. Jev selects an ordered pair
using only the input and expert descriptions; composition retains the first
update, projects the second onto the orthogonal complement of the first
update's row space, and combines them with equal weights. Across 14 PorTAL
tasks and three Qwen3 model scales, \method{} achieves the highest task-macro
and sample-micro accuracy among the evaluated methods.

\section{ACKNOWLEDGMENT}

No funding was received for this study. The authors have no relevant financial or nonfinancial interests to disclose.

\section{COMPLIANCE WITH ETHICAL STANDARDS}

No ethical approval was required for this study.
\bibliographystyle{IEEEbib}
\bibliography{refs}
\end{document}